%% file: main.tex
\documentclass[journal]{IEEEtran}

\ifCLASSINFOpdf
\else
   \usepackage[dvips]{graphicx}
\fi
\usepackage{url}
\usepackage{amsmath}
\usepackage{amssymb}
\usepackage{booktabs}
\usepackage{multirow}
\usepackage{color}
\newcommand{\figref}[1]{Fig.\ref{#1}}
\newcommand{\tabref}[1]{Tab.\ref{#1}}

\newcommand{\ie}{\textit{i}.\textit{e}.}

\newcommand{\best}[1]{\textbf{#1}}   

\newcommand{\xcy}[1]{{\color{black} {#1}}}
\newcommand{\hbz}[1]{{\color{black} {#1}}}
\usepackage{graphicx}

\begin{document}

\title{Adapting Human Mesh Recovery with Vision-Language Feedback}

\author{Chongyang Xu, Buzhen Huang, Chengfang Zhang, Ziliang Feng, Yangang Wang
\thanks{This work was supported by the Sichuan Science and Technology Program (No. 2024NSFSC2046).}
\thanks{Chongyang Xu and Ziliang Feng are with the College of Computer Science, Sichuan University, Chengdu 610065, China.}
\thanks{Buzhen Huang and Yangang Wang are with the Key Laboratory of Measurement and Control of Complex Systems of Engineering, Ministry of Education, and the
School of Automation, Southeast University, Nanjing 210096, China.
}
\thanks{Chengfang Zhang is with the Intelligent Policing Key Laboratory of Sichuan Province, Sichuan Police College, Luzhou, 646000, China.}
\thanks{Corresponding author: Yangang Wang. E-mail: yangangwang@seu.edu.cn.}
}

\maketitle



\input{secs/02_abstract}
\vspace{-8mm}

\input{secs/03_intro}
\input{secs/05_method}
\input{secs/06_experiments}
\input{secs/07_conclusion}

\bibliographystyle{IEEEbib}
\bibliography{references}

\input{supplementary}

\end{document}

%% file: secs/02_abstract.tex
\begin{abstract}

Human mesh recovery can be approached using either regression-based or optimization-based methods. Regression models achieve high pose accuracy but struggle with model-to-image alignment due to the lack of explicit 2D-3D correspondences. In contrast, optimization-based methods align 3D models to 2D observations but are prone to local minima and depth ambiguity. In this work, we leverage large vision-language models (VLMs) to generate interactive body part descriptions, which serve as implicit constraints to enhance 3D perception and limit the optimization space. Specifically, we formulate monocular human mesh recovery as a distribution adaptation task by integrating both 2D observations and language descriptions. To bridge the gap between text and 3D pose signals, we first train a text encoder and a pose VQ-VAE, aligning texts to body poses in a shared latent space using contrastive learning. Subsequently, we employ a diffusion-based framework to refine the initial parameters guided by gradients derived from both 2D observations and text descriptions. Finally, the model can produce poses with accurate 3D perception and image consistency. Experimental results on multiple benchmarks validate its effectiveness. The code will be made publicly available.


\end{abstract}

%% file: secs/03_intro.tex
\section{Introduction}
\hbz{
\IEEEPARstart{M}{ONOCULAR} human mesh recovery aims to reconstruct 3D human meshes from a single image, which can be applied to various downstream tasks, such as 3D pose estimation~\cite{splposeestimation, splposeestimation1}, person re-identification~\cite{splperson-re1, splperson-re2, splperson-re3}, and crowd analysis~\cite{splcrowd}. This task is typically addressed using either regression-based~\cite{HMR, HMR2.0} or optimization-based~\cite{SMPLify, Refit} methods. Recent regression models~(\figref{fig:head_fig}(a)) leverage extensive human data to learn pose priors, enabling the prediction of accurate joint positions and body meshes. However, they often face challenges in aligning 3D models with 2D images due to the absence of explicit 2D-3D correspondences. In contrast, optimization-based methods(\figref{fig:head_fig}(b)) provide better model-to-image alignment but are sensitive to local minima and depth ambiguity, resulting in suboptimal joint accuracy. Additionally, off-the-shelf detectors\cite{fang2022alphapose} may introduce noises, which can degrade 3D reconstruction performance.

\begin{figure}[t] 
    \centering
    \includegraphics[width=.5\textwidth]{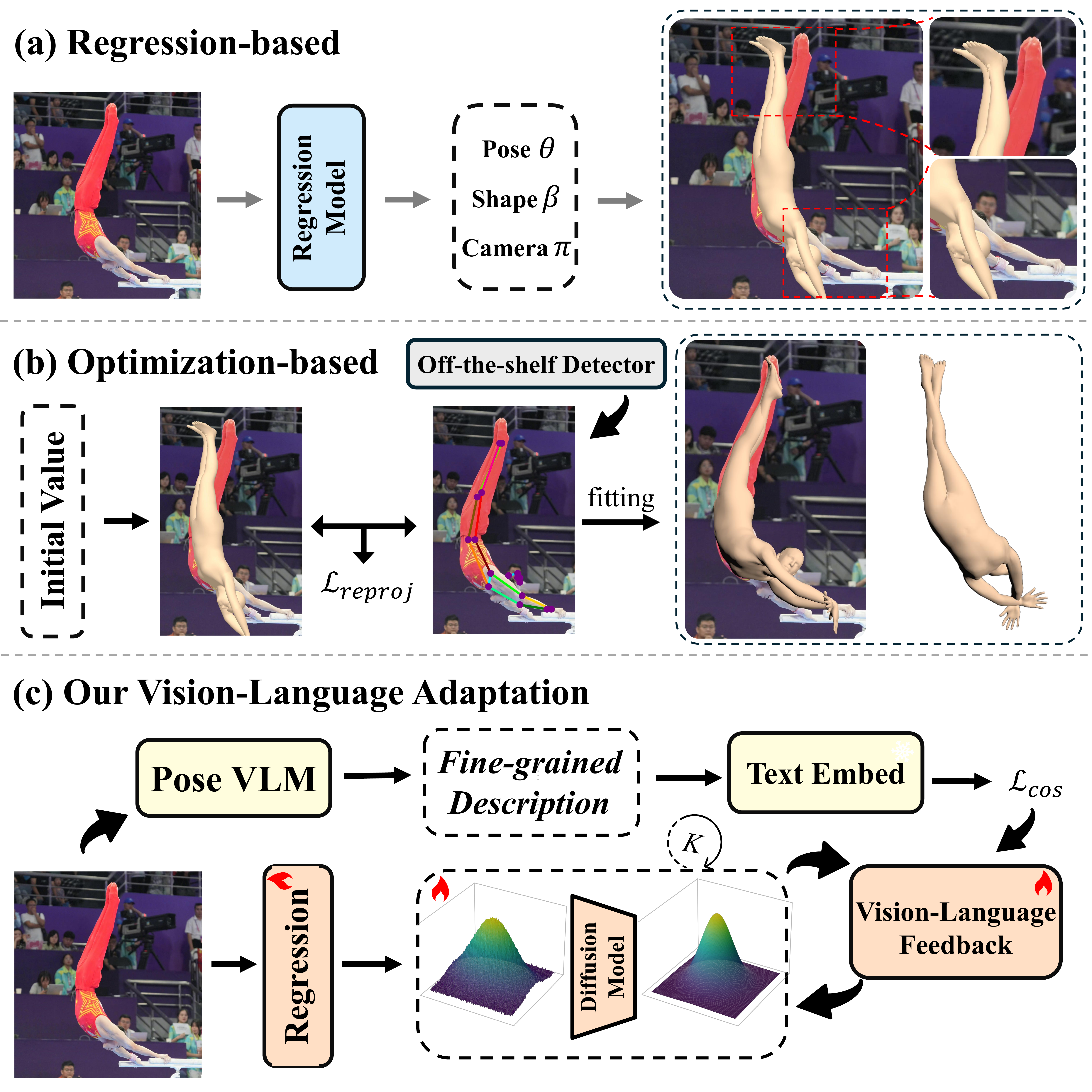}
    \caption{\hbz{(a) Regression-based methods struggle with model-image alignment for challenging poses. (b) Optimization-based methods are prone to overfitting noisy 2D inputs and suffer from severe depth ambiguity. (c) Our method leverages prior knowledge from large vision-language models to improve both 2D and 3D performance.}}
    \label{fig:head_fig}
\vspace{-8mm}
\end{figure}

Several works~\cite{kolotouros2019learning, stathopoulos2024score} have attempted to integrate regression and optimization methods into a unified framework. These approaches first train a regression model to generate initial parameters and then refine the results using additional observations, such as 2D keypoints~\cite{kolotouros2019learning} and physical laws~\cite{huang2022neural, CloseInteraction}. However, 2D keypoints are often unreliable in complex environments (e.g., occlusions). Physics-based optimization also suffers from a knowledge gap between simulation and the real world, which may result in suboptimal simulated outcomes under the given constraints. Therefore, existing approaches have yet to fully address the trade-off between image observations and model-based assumptions.

Recently, human motion generation works~\cite{Guo_2022_CVPR, petrovich23tmr, jiang2024motiongpt} reveal that texts can provide rich 3D pose information. Therefore, our key idea is to leverage textual descriptions from large vision-language models (VLMs)~\cite{ChatPose} to compensate for insufficient 2D image observations. Benefiting from the 3D reasoning capabilities of VLMs (e.g., a person sitting with one leg crossed over the other), text-image inputs can enhance 3D perception and 2D-3D consistency for human pose estimation, thereby reducing the trade-off between image observations and model-based assumptions.
}

\hbz{To this end, we propose a framework that combines regression and optimization approaches, leveraging both image observations and vision-language models (VLMs) to facilitate human mesh recovery. The initial pose is first predicted using a Vision Transformer (ViT)~\cite{ViT}, which may be inaccurate due to depth ambiguity. To refine the pose, part-aware interactive descriptions are further extracted from the image using a Vision-Language Model (VLM)~\cite{ChatPose} with carefully designed prompts. Since text cannot directly provide detailed pose information, we define the alignment between pose and text in the latent space as a guiding signal. Consequently, we train a shared space based on VQ-VAE to bridge the gap between these two modalities. In the reverse diffusion process, we evaluate the reconstructed pose using re-projection error and text-pose similarity, and then use the derived gradients as conditions in each timestep. With the text-image conditions, the initial pose is iteratively updated and will ultimately converge to the real pose. In summary, our key contributions are: (1) We propose a framework that integrates multi-modal feedback to achieve both accurate 3D pose estimation and precise model-image alignment. (2) We demonstrate that fine-grained textual interactive descriptions can enhance human mesh recovery. (3) We introduce a novel conditioning mechanism that combines vision and language observations to guide the diffusion process.}

%% file: secs/05_method.tex
\vspace{-4mm}
\section{Method}

\hbz{
In this work, we aim to reconstruct the human mesh from monocular images by optimizing body parameters to achieve accurate alignment with vision-language observations.

\vspace{-5mm}

\subsection{Preliminaries}

We use SMPL model~\cite{SMPL} with 6D representation~\cite{6D_Rotation} to represent 3D humans, and thus the parameters for a single person consists of pose $ \theta \in \mathbb{R}^{144} $, shape $ \beta \in \mathbb{R}^{10}$, and translation $\pi \in \mathbb{R}^{3}$. 

}
\vspace{-3mm}

\hbz{
\subsection{Initial Prediction}
Many diffusion-based methods~\cite{stable_diffusion} in the image generation domain rely on sampling from Gaussian noise and require numerous iterative steps during training. This results in a significant demand for large datasets and substantial computational resources, making direct image generation with diffusion models computationally expensive. To mitigate this, we follow \cite{CloseInteraction} to obtain an initial pose estimate through a regression-based approach, which serves as the starting point for the subsequent optimization process, ultimately ensuring more accurate human mesh recovery. To extract image features $I$, we use ViT~\cite{ViT} as the backbone, and integrate bounding-box information to regress the SMPL parameters, which is similar to CLIFF~\cite{li2022cliff}, where the translation $\pi$ is derived from the estimated camera parameters and transformed into the global coordinate system. The regressor network is trained on normal datasets by the following loss function:
\begin{equation}\label{5}
\mathcal{L}_{regressor} = \lambda_{smpl}\mathcal{L}_{smpl} + \lambda_{joint}\mathcal{L}_{joint} + \lambda_{reproj}\mathcal{L}_{reproj},
\end{equation}
where $\mathcal{L}_{smpl} = ||[\beta, \theta]-[\hat{\beta},  \hat{\theta}]||_2^2$, $\mathcal{L}_{joint} = ||J_{3D} - \hat{J}_{3D}||_2^2$
and the reprojection loss is given by: $\mathcal{L}_{reproj} = ||\Pi{(J_{3D})} - \hat{J}_{2D}||_2^2$,
where $\Pi(\cdot)$ projects the 3D joints to 2D image with camera parameters, and $\hat{J_{2D}}$ is the ground-truth 2D keypoints. $\lambda_{smpl}$, $\lambda_{joint}$, and $\lambda_{reproj}$ control the relative importance of each term.

\vspace{-2mm}
\hbz{
\subsection{Description Extraction and Modal Alignment}

Texts contain rich 3D information for describing human body poses, such as joint positions, part orientations and intra-body interactions, which provide essential cues to improve 3D pose estimation.
}

\vspace{-3mm}
\hbz{
\subsubsection{Description Extraction} We describe the human in the image with overall and part-based (e.g., head, arms, torso, and legs) textual descriptions, which offer a more holistic understanding of pose perception. Initially, a large language model (LLM) is used to automatically generate prompt templates for various body parts, such as \texttt{Describe the {part} interaction of the person. How are the {parts} positioned?}. Following this, the images and chosen prompts are fed into ChatPose, which generates the corresponding pose descriptions. Additional details are provided in Sup. Mat. A.
}

\begin{figure}[t] 
    \centering
    \includegraphics[width=.5\textwidth]{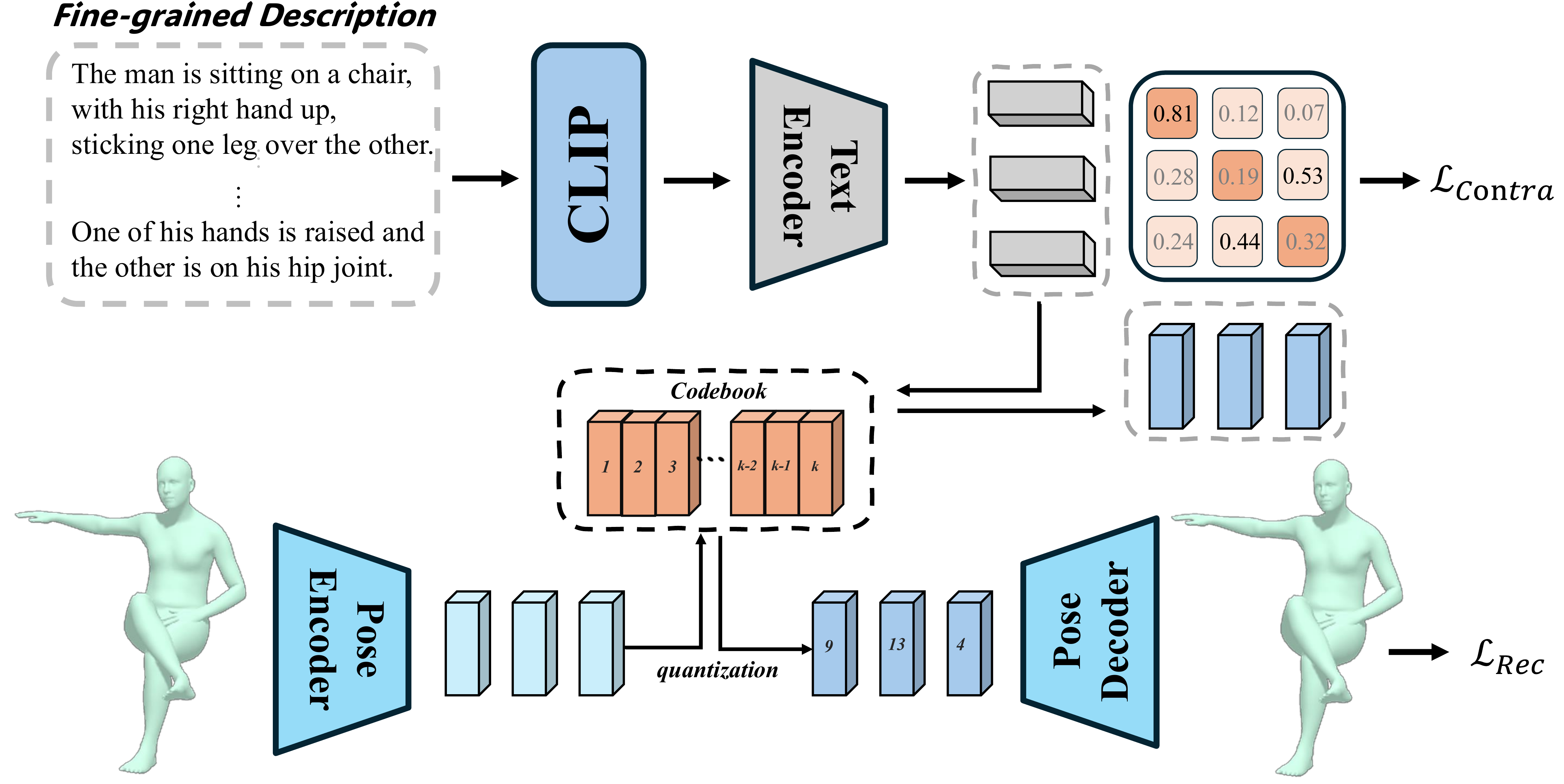}
    \caption{\textbf{Pose-Text Alignment.} \hbz{We first train a discrete pose codebook via VQ-VAE. To bridge the gap between text and 3D pose modalities, we then train a text encoder to align the texts to body poses in latent space with contrastive learning.} 
    }
    \label{fig:alignment_fig}
    \vspace{-4mm}
\end{figure}

\begin{figure*}[t] 
\centering
\includegraphics[width=\textwidth]{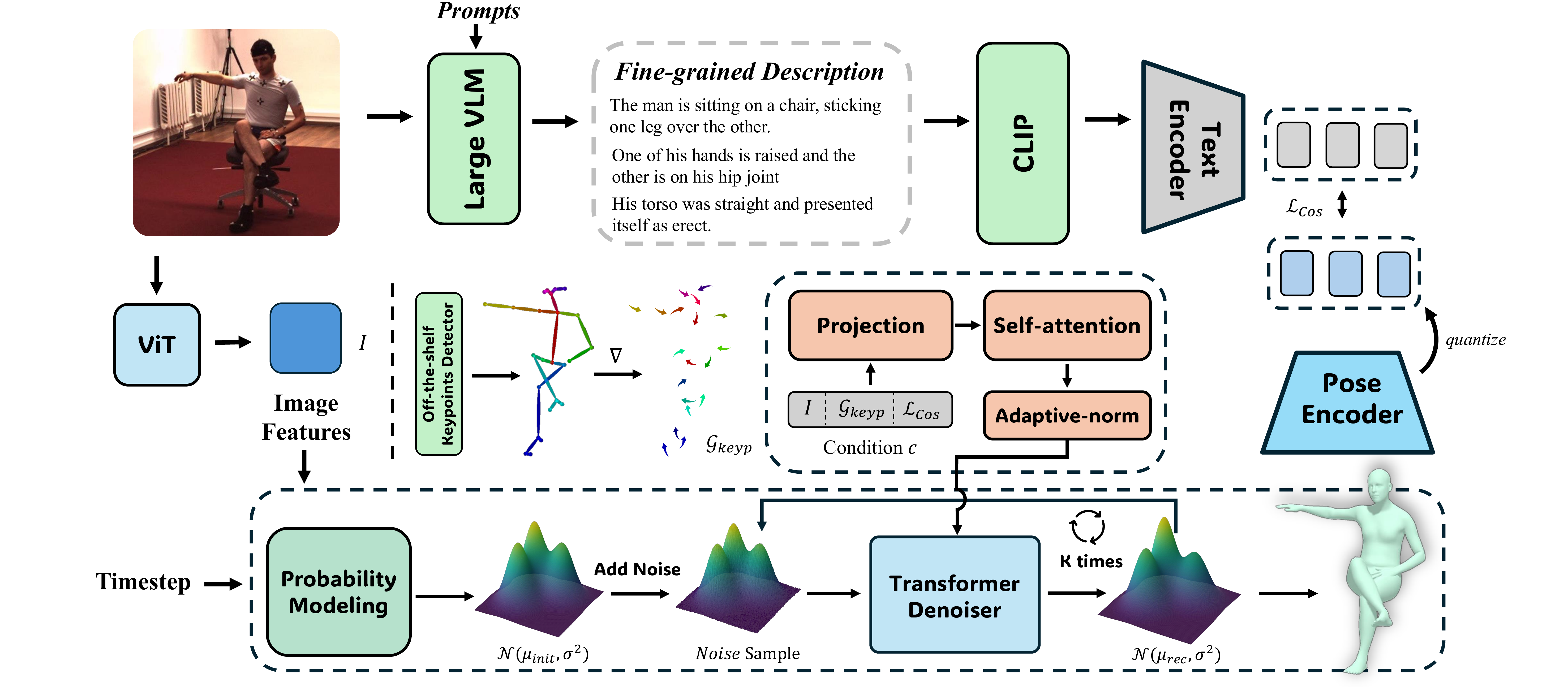}
\caption{\textbf{Overview of our method.} Given an image, a large vision-language model is first used to extract detailed interactive descriptions for the body parts. An initial prediction is then made, followed by the construction of a diffusion-based framework that refines this prediction using multi-modal feedback. At each time step, the gradients of 2D keypoints are computed, along with the similarity loss between text embeddings and the pose, while image features from the backbone are concatenated to form the condition $c$, which is then fed into the diffusion model to estimate the noise. The distribution is updated based on this guidance, ultimately yielding accurate body pose estimations.}
\label{fig:pipeline}
\vspace{-4mm}
\end{figure*}

\subsubsection{Pose-Text Alignment} \hbz{CLIP~\cite{CLIP} learns word embeddings through large-scale image-text contrastive learning, aligning representations with natural language distributions. However, these embeddings lack explicit structural information, such as joint positions and pose angles, which are essential for pose tasks. Thus, additional alignment between pose and text embeddings is necessary for pose optimization. 

For pose representation, we use VQ-VAE~\cite{van2017neural}, which quantizes the latent space into discrete encoding vectors, capturing structural features like joint positions and angles more effectively than traditional VAEs, which suffer from gradient vanishing and blurry generation. VQ-VAE's discrete representation is better suited for pose tasks as it directly encodes discrete features critical for pose. We train the VQ-VAE by optimizing the following objective:
\begin{equation}
\mathcal{L}_{vq} = \alpha \| \mathcal{E}_p(\theta) - \text{sg}[ \hat{Z} ] \|^2 + \| \mathcal{D}_p(\hat{Z}) - \theta \|^2,
\end{equation}
where $\mathcal{E}p$ and $\mathcal{D}p$ denote the pose encoder and decoder, respectively. The tokens in the codebook are represented by $\hat{Z}$. $\text{sg}[\cdot]$ and $\alpha$ refer to the stop-gradient operator and a hyperparameter. We begin by embedding the text into the CLIP space, represented as $f{c}$, and then use $\mathcal{E}t$ to map it into the pose feature space. To align the pose and text features, we apply a contrastive loss, and further refine the alignment through the reconstruction loss of the text features via the pose decoder. The following objective is used:}
\xcy{
\begin{equation}
\mathcal{L}_\text{align} = \underbrace{- \frac{1}{N} \sum_{i=1}^{N} \log \frac{\exp(z_i^{\text{pose}} \cdot z_i^{\text{text}} / \tau)}{\sum_{j=1}^{N} \exp(z_i^{\text{pose}} \cdot z_j^{\text{text}} / \tau)}
}_{\mathcal{L}_{\text{contra}}} + \mathcal{L}_{\text{rec}},
\end{equation}
\noindent where \( \mathcal{L}_{\text{rec}} = \| \mathcal{D}_p(\mathcal{E}_t(f_{c})) - \theta \|^2 \) and \( \mathcal{L}_{\text{contra}} \) represent the reconstruction and contrastive losses, respectively. \( z^{\text{pose}} \) and \( z^{\text{text}} \) represent the latent variables obtained from the encoder for the pose and text. $\tau$ is the temperature parameter, used to scale the similarity.}

\subsection{Vision-Language Feedback Adaptation}

Since the initial prediction involves minor misalignments and 3D pose errors, we formulate the optimization process as a distributional optimization, where the initial value serves as the mean of the initial distribution. We assume a probability distribution around the initial prediction, representing the potential optimization space. This allows us to fine-tune and optimize the model based on this distribution.

\subsubsection{Diffusion process} We assume that the optimized result $x$ follows a Gaussian distribution with the initial prediction \( \hat{x}^{\text{init}} \) as the mean and \( \sigma \) as the standard deviation. Assume the true distribution is $p(x)$. By training a diffusion model based on the contrast of the log gradients, \ie, $s_{\text{model}}(x; \phi) = \nabla_x \log q_\phi(x)$, the initial distribution can undergo gradient descent towards the true data distribution by the following loss function:
\begin{equation}
\mathcal{L}(\phi) = \mathbb{E}_{x \sim p(x)} \left[ \| s_{\text{model}}(x; \phi) - s_{\text{data}}(x) \|^2 \right],
\end{equation}
where \( s_{\text{model}}(x; \phi) \) and \( s_{\text{data}}(x) \) are the gradient of the model's distribution and true data with respect to \( x \). \( \mathbb{E}_{x \sim p(x)} \) denotes the expectation with respect to the data distribution \( p(x) \). During the inference phase, we can compute the gradient of the loss function with respect to the condition \( c \), and then adjust the generated sample as the following formula:

\begin{equation}
\Delta x_t = \Delta t \cdot \nabla_x \log q(x_t \mid c),
\end{equation}
where $\Delta t$ is a scaling factor, and  $\Delta x_t$ is the change in the sample $x_t$ at the current step.

\subsubsection{Vision-Language Guided Denoising} In the diffusion network, we refine the initial distribution using multi-modal observations. We treat information from different modalities as conditions 
$c$. This design allows prior information from various modalities to help for sampling out results that satisfy the condition.

\textbf{2D Keypoints.} 2D keypoint observations serve as valuable constraints due to the rich semantic information they provide. To detect keypoints, we employ an additional keypoint detector~\cite{fang2022alphapose}, followed by the computation of the gradient of the 3D joints with respect to the detected 2D keypoints:
\begin{equation}\label{10}
\mathcal{G}_{keyp} = \frac{\partial||\Pi(J_{3D})- p_{2D}||_2^2}{\partial J_{3D}},
\end{equation}
where $p_{2D}$ represents the detected 2D keypoints, and $J_{3D}$ is the set of 3D joints, computed as a linear combination of vertices: $J=WM$.

\textbf{Text.} Since human pose and depth are strongly coupled, relying solely on 2D information often leads to poor performance due to depth ambiguity and information loss after projection; thus, we consider text as additional information to implicitly constrain the body pose in 3D space. Specifically, we compute the similarity loss between the pose and text features in the latent space:
\begin{equation}
\mathcal{L}_{cos}  =  \left(
\frac{\mathcal{E}_p(\theta)\cdot\mathcal{E}_t(f_{c})}{\|\mathcal{E}_p(\theta)\|\|\mathcal{E}_t(f_{c})\|} \right)^2,
\end{equation}
The gradient of the pose parameters with respect to the similarity loss, $\mathcal{G}_{text} = \frac{\partial \mathcal{L}_{cos}}{\partial \hat{\theta}}$, can implicitly provide guidance. Finally, the condition $c = \text{concat}(I,\mathcal{G}_{\text{keyp}}, \mathcal{G}_{\text{text}})$ serves as vision-language feedback to optimize the sampling process.}

\vspace{-2mm}

%% file: secs/06_experiments.tex
\section{Experiments}
\xcy{
\subsection{Experimental setup}

\subsubsection{Datasets and Metrics.} In line with previous studies, we utilize the Human3.6M~\cite{Human3.6m}, COCO~\cite{COCO}, MPII~\cite{MPII}, and MPI-INF-3DHP~\cite{3DHP} datasets for training. These image and video datasets are employed to train both the regressor network and the diffusion model. We evaluate our method on the 3DPW test split~\cite{3DPW} and the Human3.6M validation split~\cite{Human3.6m}. For 3D pose accuracy, we report the Mean Per Joint Position Error (MPJPE), as well as the MPJPE after rigid alignment of the predicted poses with the ground truth (PA-MPJPE).

\subsubsection{Implementation Details.} First, we train the initial prediction regressor. We adopt the ViT-H/16~\cite{ViT} and the standard transformer decoder~\cite{transformer}, as proposed in~\cite{HMR2.0}. We use ChatPose~\cite{ChatPose} to extract descriptive information, and AlphaPose~\cite{fang2022alphapose} as an additional keypoint detector to provide 2D keypoint data. Next, we align the pose and text features in the latent space. Finally, we train the diffusion optimization module while keeping the other modules frozen. For training the regressor, we use 20 epochs with a batch size of 128 and a learning rate of $1e{-5}$. The pose-text alignment is performed across multiple datasets for 100 epochs with a batch size of 256. During diffusion training, we run 30 epochs with a batch size of 128 on four RTX 3090 GPUs.

\vspace{-4mm}

\subsection{Comparisons with the state-of-art methods}

\begin{figure}[t] 
\centering
\includegraphics[width=0.5\textwidth]{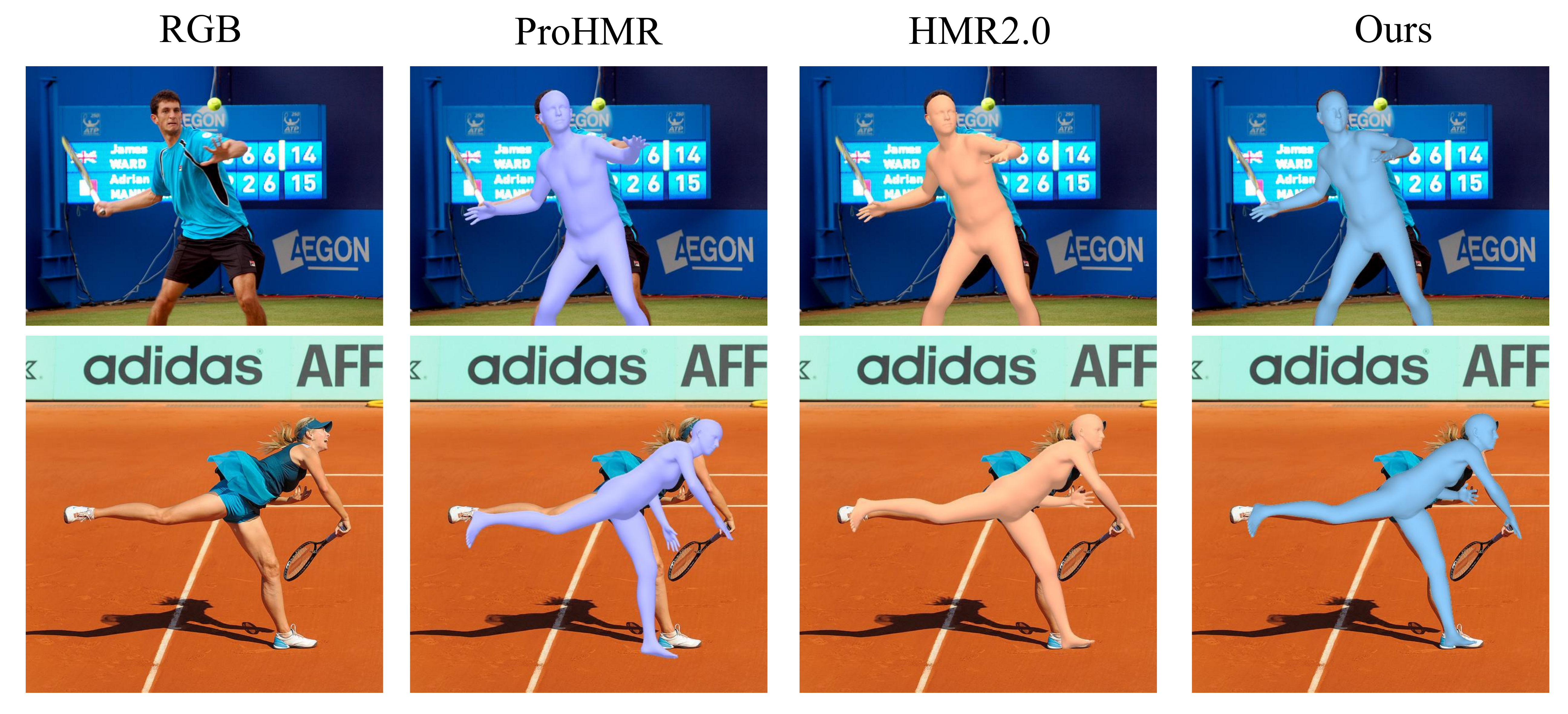}
\caption{\textbf{Qualitative results.} From left to right: RGB image, ProHMR~\cite{kolotouros2021probabilistic}, HMR2.0~\cite{HMR2.0}, and our method. Our approach ensures accurate 3D joint positions with minimal depth ambiguity while achieving robust front-facing alignment.}
\label{fig:quantative}
\end{figure}

We compare our method with state-of-the-art human mesh recovery approaches on the Human3.6M and 3DPW datasets, reporting MPJPE and PA-MPJPE metrics in \tabref{tab:3d_metrics}. Since our method incorporates multiple conditional constraints, it outperforms most existing methods. Specifically, the PA-MPJPE improves by 0.4 on the 3DPW dataset and by 1.2 on the Human3.6M dataset. We also present a qualitative comparison in \figref{fig:quantative}. While HMR2.0 exhibits deviations in mesh alignment, ProHMR faces challenges, particularly in cases of depth ambiguity, leading to poorer optimization results. In contrast, our method demonstrates enhanced robustness, as the textual information provides supplementary context that helps mitigate the limitations of unreliable 2D observations.

\begin{table}[t]
  \centering
   \caption{\textbf{Reconstruction evaluation on 3D joint accuracy.} We report reconstruction errors on the 3DPW and Human3.6M datasets. 
    }
  \footnotesize
    \begin{tabular}{l|cc|cc}
    \toprule[1pt]
    \multicolumn{1}{c|}{\multirow{2}{*}{Method}} & \multicolumn{2}{c|}{3DPW} & \multicolumn{2}{c}{Human3.6M} \\
         & MPJPE & PA-MPJPE & MPJPE & PA-MPJPE \\
         \midrule[1pt]
        HMR~\cite{kanazawa2018end} & 130.0 & 76.7 & 88.0 & 56.8 \\
        SPIN~\cite{kolotouros2019learning} & 96.9  & 59.2 & 62.5  & 41.1 \\
        DaNet~\cite{zhang2019danet} &  -  &  56.9   & 61.5  & 48.6 \\
        PyMAF~\cite{zhang2021pymaf}               & 92.8        & 58.9         & 57.7         & 40.5         \\
        ProHMR~\cite{kolotouros2021probabilistic} & - & 55.1 & - & 39.3 \\
        PARE~\cite{kocabas2021pare}                & 82.0& 50.9 & 76.8         & 50.6         \\
        PyMAF-X~\cite{zhang2023pymaf}             & 78.0& 47.1 & 54.2 & 37.2 \\
        HMR 2.0~\cite{HMR2.0}      & 70.0         & 44.5  & \best{44.8}  & 33.6  \\
        \textbf{Ours} & \best{69.3} & \best{43.9} & 47.7 & \best{32.4} \\
    \bottomrule[1pt]
    \end{tabular}%
  \label{tab:3d_metrics}%
  
\end{table}%

\vspace{-3mm}

\subsection{Ablation study}

\begin{table}[t]
\centering
\footnotesize
\caption{\textbf{Ablation study.} The initial parameters are regressed by the regressor. We report the results under different conditions in the diffusion process. All numbers are in millimeters (mm).}
\begin{tabular}{l|cc|cc}
    \toprule[1pt]
    \multirow{2}{*}{Method} & \multicolumn{2}{c|}{3DPW} & \multicolumn{2}{c}{Human3.6M} \\
                  & MPJPE & PA-MPJPE & MPJPE & PA-MPJPE \\
    \midrule[1pt]
    Standard Gaussian  & 87.8 & 54.9 & 62.8 & 44.6  \\
    Initial Prediction & 73.4 & 47.5 & 56.4 & 34.0  \\
    \ w/ image      &  70.6   & 45.6  &  53.5  &  33.4   \\
    \ w/ keypoints  &  72.3   & 46.0  &  54.4  &  33.7\\
    \ w/ text       &  72.8   & 47.0  &  56.2  &  33.9 \\
    \ w/o keypoints &  70.3   & 45.1  &  52.7  &  33.1 \\
    \ w/o text      &  69.8   & 44.5  &  48.3  &  32.8 \\
    \ w/ all conditions &  \best{69.3}   & \best{43.9}  &  \best{47.7}  &  \best{32.4} \\
    
    \bottomrule[1pt]
\end{tabular}
\vspace{-3mm}
\label{table2}
\end{table}

\subsubsection{Initial Prediction.}
We investigated the importance of the initial regressor and found that, compared to a standard Gaussian distribution, using one with prior knowledge of human pose leads to better optimization results.
\subsubsection{Multi-modal Conditions}
We further investigate the impact of different conditions during the optimization process. We report the results for three scenarios: using a single modality condition (denoted as "w/ modality"), using all conditions except one modality (denoted as "w/o modality"), and using all conditions for optimization. Our findings show that the diffusion adaptation process effectively enhances the accuracy of initial predictions, achieving the best results when all three modalities are used. The most significant improvement comes from the image features and keypoint information, while the inclusion of text information further refines pose optimization. Text information provides additional constraints, helping to guide the optimization process and preventing it from getting stuck in local minima caused by noisy 2D keypoints.

\vspace{-1mm}

}

%% file: secs/07_conclusion.tex
\vspace{-3mm}
\hbz{
\section{Conclusion}
In this work, we propose a diffusion-based framework that combines Vision-Language Models (VLMs) and image observations for accurate human mesh recovery. By aligning pose and text within a shared latent space, we incorporate text-pose prior knowledge from VLMs. Using the diffusion model's guidance mechanism, our approach balances image observations and model assumptions through multi-modal feedback, ultimately producing body poses with precise image-model alignment and accurate joint positions after denoising.
}

%% file: supplementary.tex
\clearpage

\twocolumn[
\begin{center}
    \fontsize{16}{25}\selectfont \textbf{Adapting Human Mesh Recovery with Vision-Language Feedback} \par
    \textnormal{Supplementary Material}
\end{center}
]

In this supplementary material, we provide additional details on data processing, model architecture, and more qualitative results.

\section{Additional Data Details}

\subsection{Prompt Generation} 
\begin{figure}[h]
    \centering
    \includegraphics[width=.35\textwidth]{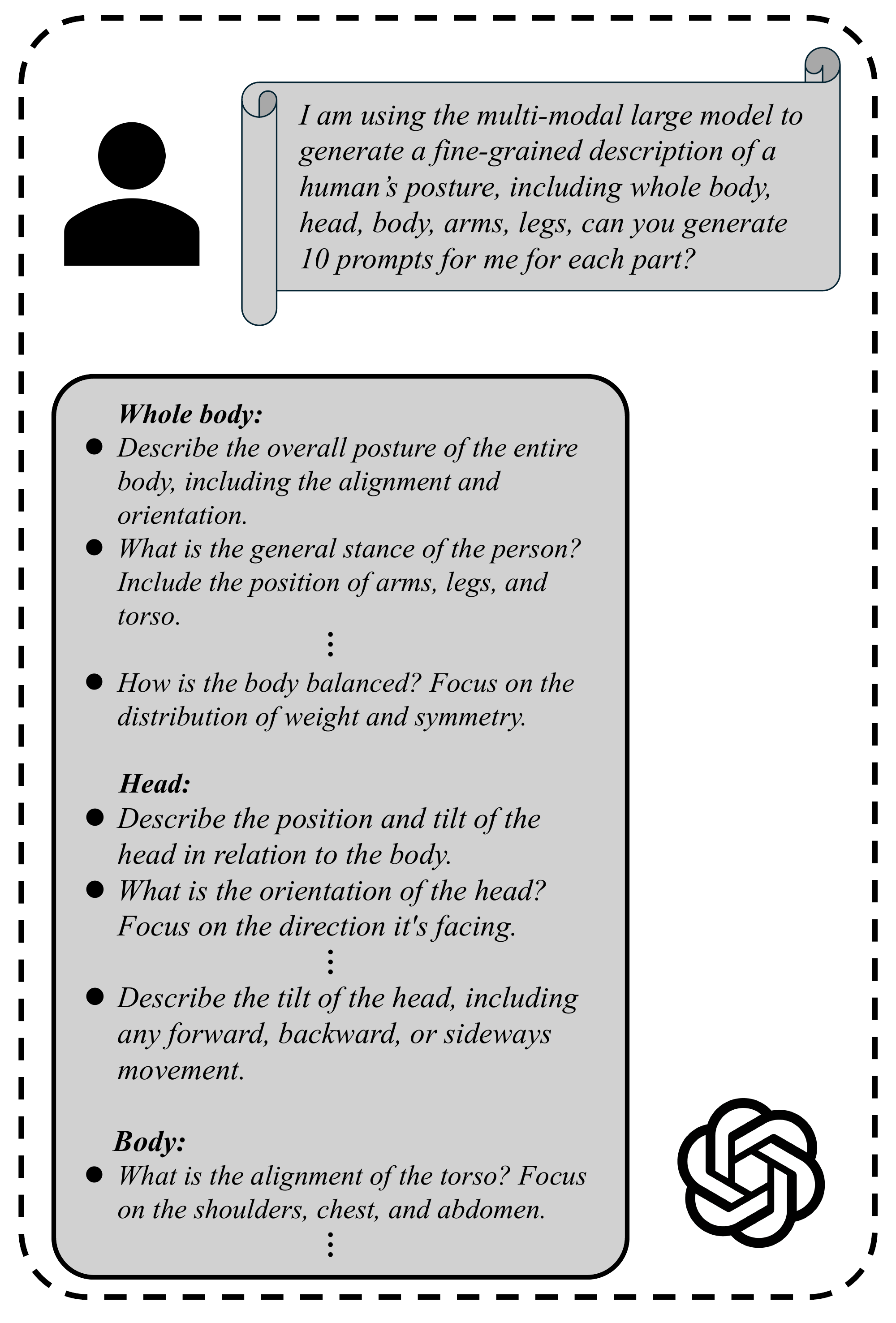}
    \caption{\textbf{Prompt Generation.} We use GPT-4~\cite{GPT4} to automatically generate 10 prompts, which are then verified and used with ChatPose~\cite{ChatPose} to describe each part of the human pose.}
    \label{fig:sup_gpt}
\end{figure}

Prompt engineering plays a crucial role in enhancing both performance and efficiency. Given the complexity of accurately describing body poses, carefully crafted prompts are used to extract detailed human pose descriptions. Specifically, GPT-4~\cite{GPT4} is first employed to automatically generate ten prompt sentences for each body part, as shown in Figure~\ref{fig:sup_gpt}. These prompts are then manually reviewed to ensure their accuracy and minimize ambiguity.

\vspace{-4mm}

\subsection{Part Description Generation.} 
\begin{figure}[h]
    \centering
    \includegraphics[width=.40\textwidth]{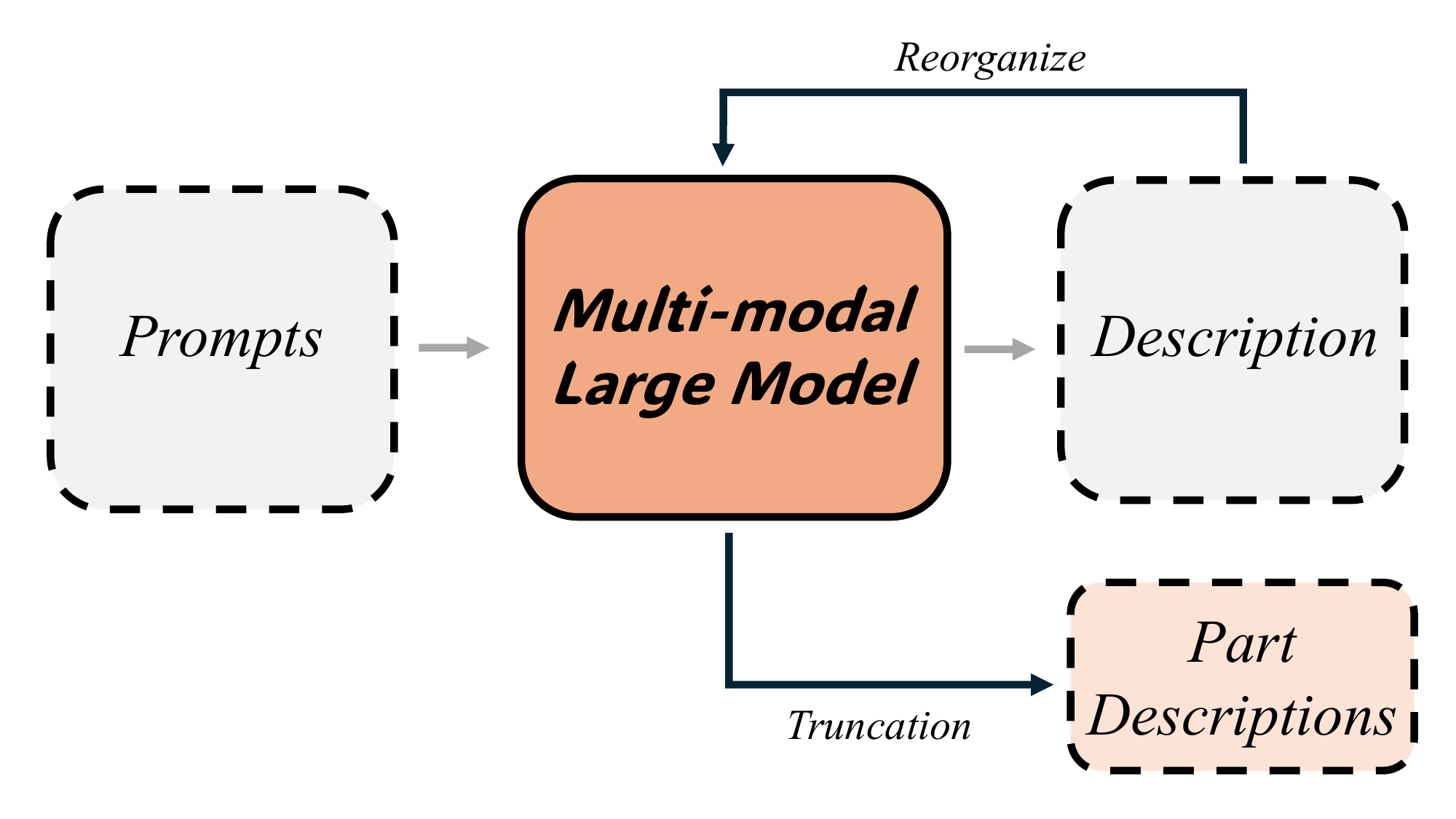}
    \caption{\textbf{Schematic of description generation and reorganization.} We use the generated prompts to create descriptive texts for each image. The key information from these texts is then extracted, and the final description is truncated to 77 words or fewer.}
    \label{fig:sup_des_process}
\end{figure}

Using the generated prompts, we feed the RGB image, human bounding box, and prompt sentences for each body part into ChatPose~\cite{ChatPose}, an open-source large model designed for extracting pose descriptions, as shown in Figure~\ref{fig:sup_des_process}. Since the character in video datasets exhibits minimal movement over short periods, we extract text descriptions from the 30th frame of every 60-frame sequence. Since CLIP~\cite{CLIP} accepts only short sentences, we use ChatPose to reorganize the key information and limit the final description to 77 words or fewer.

\begin{figure}[]
    \centering
    \includegraphics[width=.36\textwidth]{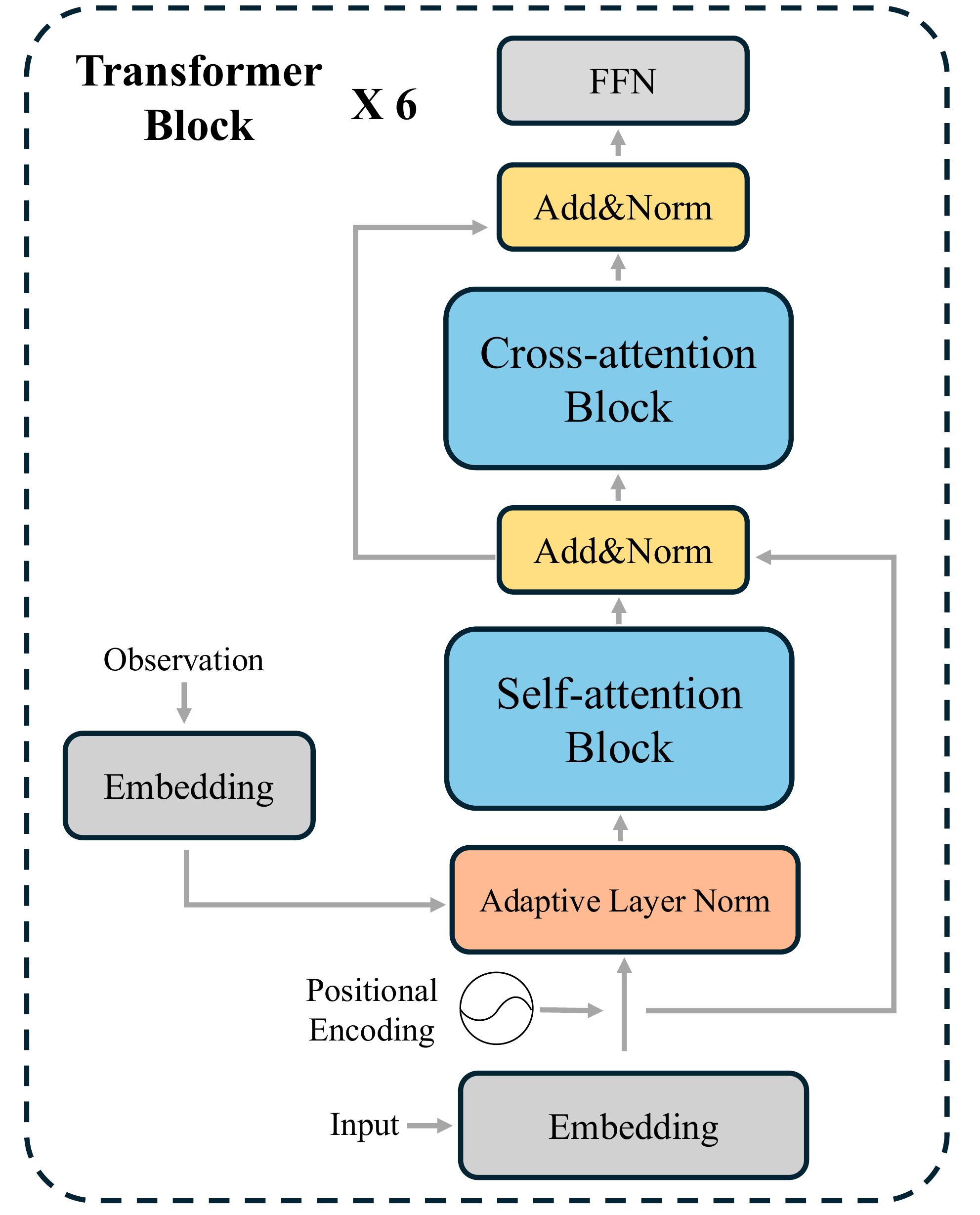}
    \caption{\textbf{Architecture of our conditional diffusion model.} We adopt the Transformer architecture and replace the standard normalization layer with an adaptive normalization layer. This layer combines the noisy SMPL parameters, positional embeddings, and observations through adaptive normalization.}
    \label{fig:sup_denoiser_network}
\end{figure}

\begin{figure}
    \centering
    \includegraphics[width=\linewidth]{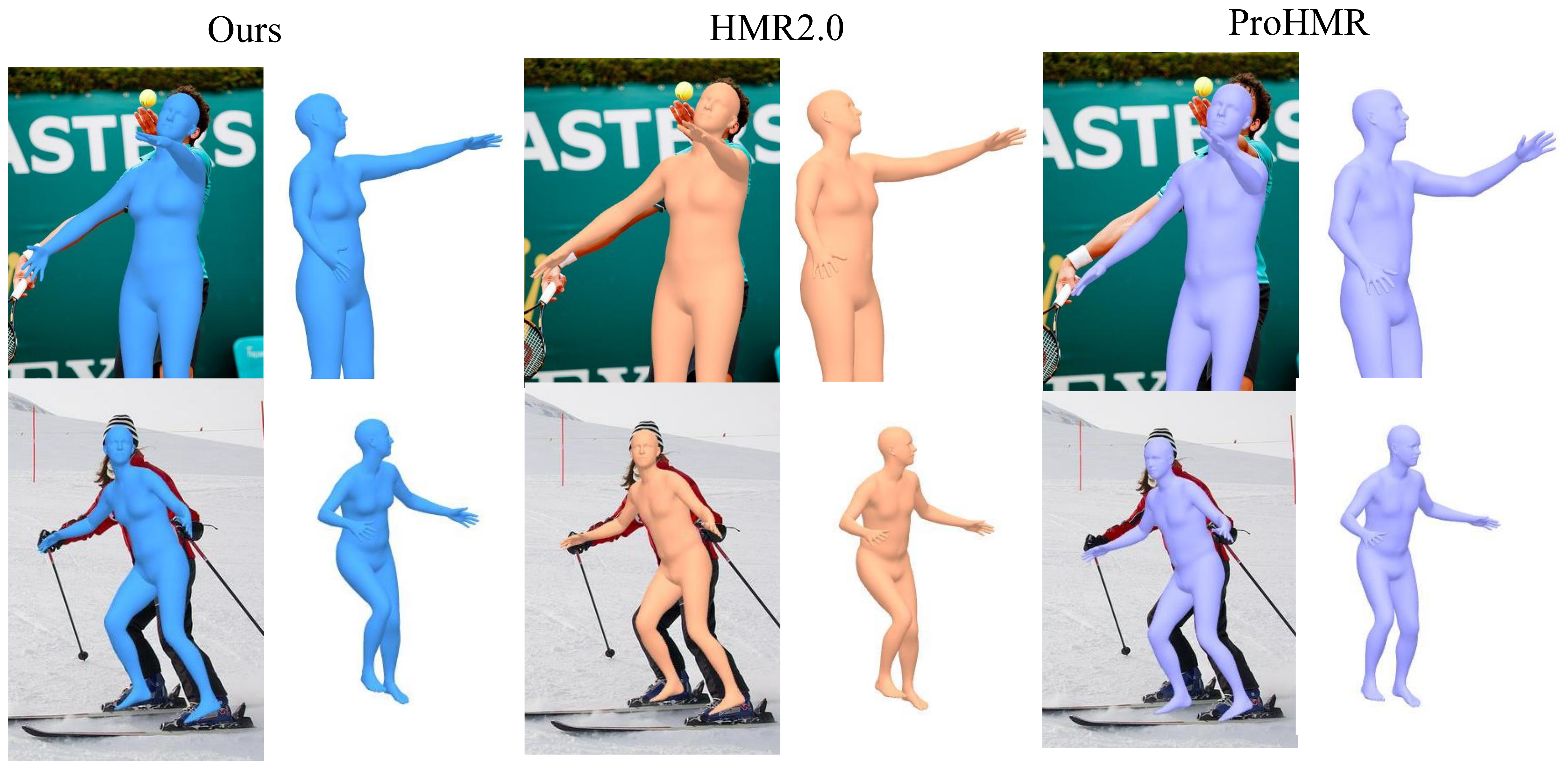}
    \caption{\textbf{More qualitative results.} From left to right are our method, HMR2, and ProHMR, including both front and side views. Our method has good alignment with better 3D accuracy.}
    \label{fig:enter-label}
\end{figure}